arXiv:cs/0602051v1 [cs.NE] 14 Feb 2006# The University of Algarve
# Informatics Laboratory

UALG-ILAB
Technical Report No. 200601
February, 2006**On the utility of the multimodal problem generator for assessing the performance of Evolutionary Algorithms**

**Fernando G. Lobo**, and **Cláudio F. Lima**Department of Electronics and Informatics Engineering
Faculty of Science and Technology
University of Algarve
Campus de Gambelas
8000-117 Faro, Portugal
URL: \protect\vrulewidth0pthttp://www.ilab.ualg.pt
Phone: (+351) 289-800900
Fax: (+351) +351 289 800 002

# On the utility of the multimodal problem generator for assessing the performance of Evolutionary Algorithms

**Fernando G. Lobo** *
UAlg Informatics Lab
DEEI-FCT, University of Algarve
Campus de Gambelas
8000-117 Faro, Portugal
flobo@ualg.pt

**Cláudio F. Lima**
UAlg Informatics Lab
DEEI-FCT, University of Algarve
Campus de Gambelas
8000-117 Faro, Portugal
clima@ualg.pt

**Abstract**

This paper looks in detail at how an evolutionary algorithm attempts to solve instances from the multimodal problem generator. The paper shows that in order to consistently reach the global optimum, an evolutionary algorithm requires a population size that should grow at least linearly with the number of peaks. It is also shown a close relationship between the supply and decision making issues that have been identified previously in the context of population sizing models for additively decomposable problems.

The most important result of the paper, however, is that solving an instance of the multimodal problem generator is like solving a *peak-in-a-haystack*, and it is argued that evolutionary algorithms are not the best algorithms for such a task. Finally, and as opposed to what several researchers have been doing, it is our strong belief that the multimodal problem generator is not adequate for assessing the performance of evolutionary algorithms.

## 1 Introduction

The multimodal problem generator has first been introduced by De Jong, Potter, and Spears (1997). The generator creates problem instances with a controllable number of peaks. The major motivation for developing that (and other) test problem generators is primarily connected to a lack of a sound methodology for conducting experimental research. For many years, researchers have been proposing new algorithms, or variations on existing algorithms, and often assess their performance on a limited number of ad-hoc test functions hoping that the results can be generalized beyond that. Oftentimes, the results do not generalize beyond those functions, and so, there is a large number of research papers with unfair comparisons and claims that lack substantial support.

In addition to that, it is well known that it is not possible to have a search algorithm that is better than another search algorithm for all possible problems (Wolpert & Macready, 1997). The no-free-lunch argument, however, is usually not a big concern for researchers in the field because it is widely accepted that evolutionary algorithms (EAs) do not attempt to solve all (random) problems, but rather attempt to solve problems that have some sort of structure that can be exploited by the EA's internal mechanisms.

Because of the above reasons, EA researchers have been trying to identify classes of problems in order to find appropriate algorithms for solving instances of those classes. The idea of using a test problem generator becomes then interesting as it removes to a large degree the possibility of tweaking a particular algorithm to a particular test function.

---

*Also a member of IMAR - Centro de Modelação Ecológica.

In this paper we analyze in detail the multimodal problem generator. After its introduction in 1997, Spears used the generator in the context of assessing the usefulness of crossover versus mutation on that particular class of problems (Spears, 2002). Since then, this problem generator has been used by many researchers to assess the performance of evolutionary algorithms (Kennedy & Spears, 1998; Ochoa, Harvey, & Buxton, 1999; Alba & Troya, 2000; Hill & O'Riordan, 2003; Eiben, Marchiori, & Valko, 2004).

The rest of this text is organized as follows. We start by describing the multimodal problem generator. Then, in section 3 we set forth an hypothesis of how an EA attempts to solve any given instance of the problem generator. Section 4 is an experimental section that confirms our hypothesis. Finally, section 5 show that EAs are not the most adequate algorithms for solving instances of the multimodal problem generator. The paper is summarized in section 6, and the major conclusions are highlighted in section 7.

## 2 The multimodal problem generator

The generator creates problem instances with a certain number of peaks (the degree of multi-modality). For a problem with $\mathcal{P}$ peaks, $\mathcal{P}$ bit strings of length $L$ are randomly generated. Each of these strings is a peak (a local optima) in the landscape. Different heights can be assigned to different peaks based on various schemes (equal height, linear, logarithm-based, and so on). To evaluate an arbitrary individual $\bar{x}$, first locate the nearest peak in Hamming space, call it $Peak_n(\bar{x})$.

$$Hamming(\bar{x}, Peak_n(\bar{x})) = \min_{i=1}^{\mathcal{P}}(Hamming(\bar{x}, Peak_i))$$

Then the fitness of $\bar{x}$ is the number of bits the string has in common with that nearest peak, divided by $L$, and scaled by the height of the nearest peak. In case there is a tie when finding the nearest peak, the highest neighboring peak is chosen.

$$f(\bar{x}) = \frac{L - Hamming(\bar{x}, Peak_n(\bar{x}))}{L} \cdot Height(Peak_n(\bar{x}))$$

In this paper, and without loss of generality, we are going to assume a linear scheme for assigning heights to peaks. Under this scheme, the peak heights are linearly interpolated between a maximum value of 1.0 and a minimum value of $h < 1.0$. For instance, on a 6-peak problem with $h = 0.5$, the heights of the peaks would be 0.5, 0.6, 0.7, 0.8, 0.9, and 1.0.

On this class of problems, any string has a fitness value that range from 0.0 to 1.0. The goal of a search algorithm when optimizing an instance of this class of problems is to find the highest peak, a string with fitness 1.0. The difficulty of the problem depends on the following aspects:

- Number of peaks.
- Distribution of peak heights.
- Distribution of peaks.

The higher the number of peaks, the more difficult the problem is. Likewise, the more peaks have heights close to the global optimum, the more difficult the problem is. Finally, the location of the peaks in the search space can also make some problem instances more difficult than others. To see why this is so, consider a two-peak problem where the location of peaks are the strings 111…111 and 000…000. Without loss of generality, let us assume that the 000…000 string is the global optimum. The peaks are maximally away from each other in hamming space, and therefore, it will be very difficult for a solution that is very close to one of the peaks to move to a solution that is close to the other peak. The case when the optima are maximally away is of course an extreme situation. Another extreme situation is the case where the location of peaks are very close in hamming distance. For example, a string with all 0s, and a string with just one 1 and the rest all 0s. Under such a layout, the problem is much easier to solve because solutions can easily jump from peak to peak.

In order to wash away the effects of the layout of peaks, researchers often conduct experiments by repeating their simulations for a number of different random layout of peaks and then average the results. By doing so, the possibility that a particular layout favors or not the EA diminishes. In any case, even for a single random layout of peaks, the extreme (and close to extreme) situations are very unlikely to occur because the hamming distance between any two peaks follows a binomial distribution with $L$ Bernoulli independent trials, each with success probability $1/2$. Thus, the average distance between peaks is $L/2$, and the standard deviation is $\sqrt{L/4}$.

## 3 How an evolutionary algorithm solves these problems?

Spears (2002) has made several controlled experiments with the multimodal problem generator. The experiments included problems with peaks with equal heights as well as peaks with unequal heights using a linear interpolation as described in our previous section. The major conclusions from his experimental study was that an EA with recombination alone outperformed an EA with mutation alone on a one-peak problem, but as the number of peaks increase, recombination ends up having a deleterious effect. As opposed to that, the performance of mutation does not seem to be much affected by the number of peaks. Spears also observed that crossover benefits when the peaks have unequal heights, especially as the height $h$ of the lowest peak is reduced. In his experiments, Spears used a standard generational GA with population size of 100 individuals, fitness-proportional selection with scaling, one-point crossover with probability $P_c = 0.6$, and a bit-flip mutation rate $P_m = 0.001$. The runs were stopped after the completion of 30 thousand fitness function evaluations. Spears also observed that using other crossover operators (two-point or uniform) yielded similar results.

In the rest of this paper we are going to argue that the fundamental issue here is not between the usefulness of crossover versus mutation and vice-versa. Herein, we argue that both operators are qualitatively equally good (or equally bad) for reliably solving this class of problems. Reliability should be emphasized, as we are particularly interested in the ability of an EA to consistently find the best peak. In order to so, it doesn't really matter if recombination is better than mutation or vice-versa. As we are about to see, the critical issue is the population size. If the population size is not large enough, an EA will only be able to find the best peak occasionally. On the contrary, if the population is sized properly, an EA can reliably find the best peak.

Before getting deeper into this issue, let us think about how an EA tries to solve a problem with multiple peaks. Notice that a single-peak problem is precisely the exact same problem as the counting ones problem (often referred to as *onemax*). In the onemax problem, the fitness of a bit-string is the number of bits with value 1. Normalizing the fitness value (dividing by $L$) in onemax corresponds to a single-peak problem where the location of the peak is the string 111…111. The onemax problem has been studied extensively both in theoretical and empirical terms in the evolutionary computation literature. It is a problem that is considered easy for EAs, and that can be solved to optimality in $O(L\log L)$ fitness function evaluations, either with a crossover-based EA (Harik, Cantú-Paz, Goldberg, & Miller, 1999; Mühlenbein & Schlierkamp-Voosen, 1993) or with a mutation-based EA (Mühlenbein, 1992).

Before analyzing how an EA solves a general $\mathcal{P}$-peak problem, let us first consider the 2-peak problem described previously, where the location of the two peaks are 111…111 and 000…000. In this problem, the best solutions are strings with a lot of 1s and strings with a lot of 0s. Strings that have roughly half 0s and half 1s have low fitness values. As Spears has pointed out, crossover performs poorly when combining two solutions that are close to the top of two distinct peaks. Crossing a string with a lot of 1s with a string with a lot of 0s, is likely to yield strings in the valley between the two peaks. Notice however, that if we cross two strings near the same peak, the resulting offspring are likely to be near that same peak also. Mutation on the other hand, does a small perturbation on a solution. If we mutate a solution that is close to the top of some peak, the resulting solution will also be close to the top of that same peak, perhaps a little bit closer or a little bit further away.

In a population-based EA, the selection operator favors more fit solutions. When having multiple peaks, what is likely to occur in a regular EA (an EA without diversity preservation techniques such as niching) is that fairly quickly, the population will be concentrated in the basin of attraction of a single peak. Once that happens it really does not make much difference whether crossover or mutation is better. Once the population is focused around a single peak, the problem becomes akin to onemax and any operator (together with selection) will be able to climb up the peak.

Notice also that once the population is focused around a peak, there is no information about the search space to help to guide the algorithm to reach another peak. In other words, there is no structure to exploit. Crossover is only effective in these problems when it combines two solutions near the same peak. Otherwise it cannot do much because there is no sub-solutions to combine. The reason why there is no structure to exploit is due to the random location of peaks. Being near one peak, tell us nothing regarding as to whether other peaks might be.

For an EA to reach the top of the highest peak it has to focus its attention on that very best peak right from the early stages of the search. Herein, we argue that an EA with a small population size can only do so due to luck (for a large number of peaks, of course). To reliably solve the problem, the population sizing requirements are quite large as the GA needs to have in the initial population enough samples of solutions at the basin of attraction of the best peak, and then hope that the selection operator is able to focus the population on that region. Notice that it is not enough to have a single instance of a solution near the best peak, because in a single competition, that solution might loose with a solution near another peak. The selection operator has to reliably distinguish between solutions at the basin of attraction of the best peak, and solutions at the basin of attraction of the other peaks. These issues are analogous to the ones identified in previous population sizing models regarding building block supply and decision making (Harik, Cantú-Paz, Goldberg, & Miller, 1999; Goldberg, Deb, & Clark, 1992). The difference is that in those studies, the models are centered around the notion of a building block. In this case, however, there are no building blocks. The competition is taking place at the level of complete strings, with the different peaks competing with each other to absorb the population.

## 4 Experimental verification

To confirm our hypothesis, we run a simple genetic algorithm on a problem instance with multiple peaks, and monitor the distribution of the population members among the various peaks during the course of the run. Recall that in order to compute the fitness of a solution, we have to locate the nearest peak to that solution. In addition to computing the fitness value for the solution, we keep track of the nearest peak to that solution, and we do so for all the population members. By doing that, we can have an idea of how many solutions are at the basin of attraction of a particular peak at any given point in time.

For a problem with $\mathcal{P}$ peaks, and assuming that the random layout of peaks does not particularly favors one peak over the other, we should expect that in a randomly initialized population of size $N$, the number of solutions at the basin of attraction of a given peak is on average $N/\mathcal{P}$. In practice we might observe that the distribution is not completely uniform, but has a slight preference to better peaks. The reason for that lies in the fact that when 2 solutions are equally distant to a peak, the highest peak is considered to be the closest one.

We now show the results of running a standard GA on a 100-bit problem instance with 2, 10, and 100 peaks. The GA is generational, using two-point crossover with probability $P_c = 0.7$, binary tournament selection without replacement, bit-flip mutation with probability $P_m = 1/L = 0.01$, and population size 100. We let the GA run until it either reaches the global optimum or if a maximum of 30 thousand function evaluations have elapsed. For all problems, peak 0 is the highest peak, peak 1 is the second highest peak, peak 2 is the third highest peak, and so on. It should be stressed at this point that we are not trying to tweak the GA by any means, or that these parameters and operators are the best for these problems. What we are particularly interested in showing with the experiments is the evolution of the distribution of solutions around the peaks, and that turns out to exhibit a similar behavior for a large combination of parameter settings and operators.

Figure 1-a shows the distribution of solutions around the peaks for the first 70 generations of a two-peak problem. The results were obtained from a single run on a single problem instance. Although it is usually important to do a number of independent simulations when testing EAs, we believe that the presentation is easier by looking in detail at what happens in a single run (we did perform more independent runs and obtained similar results). Out of a total of 100 individuals, 58 were near peak 0 and 42 were near peak 1 at generation 0. The slight preference towards peak 0 can be explained in part due to chance variation alone (recall that this is a single run), and in part due to the fact that ties are resolved by assigning the nearest peak to be the highest one. Notice that fairly quickly the whole population is focused around a single peak, which happen to be the best one in this case. By generation 6, the whole population is at the basin of attraction of the best peak, and from that point on until the end of the search, the distribution of solutions around the peaks remain like that.

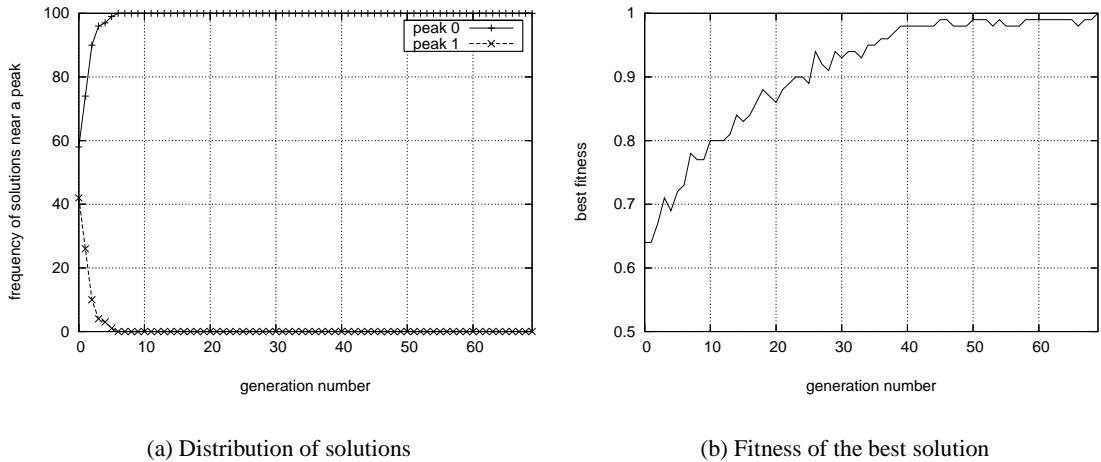

(a) Distribution of solutions    (b) Fitness of the best solution

Figure 1: Distribution of solutions around the peaks and the fitness of the best solution in the population for a 2-peak problem using a population size of 100.

Figure 1-b shows the fitness of the best solution in the population at any given point in time for the exact same run on the 2-peak problem. The interesting part to notice from both plots is that around generation 6, the population only contain solutions near peak 0, and by that time the best solution of the population is only worth 0.73 (it is at a hamming distance of 27 bits to the global optimum). But from that point on, the problem became as easy as the onemax problem, and any operator (with more or less effort) together with selection is able to climb up the peak. Notice that for these problems, the GA would be able to climb the peak faster had it used uniform crossover rather than two-point crossover, for the exact same reason why uniform crossover is also faster when solving the onemax problem. Nevertheless, that's not the point that

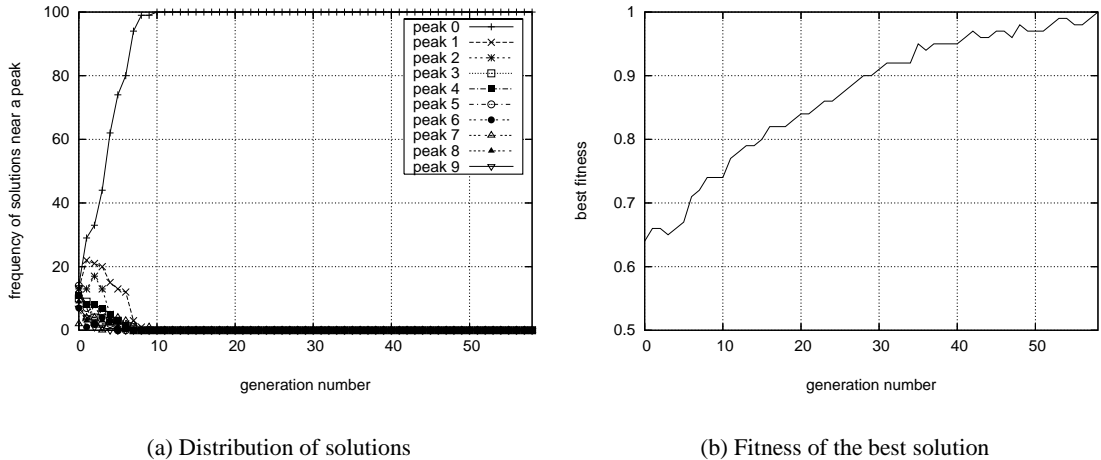

(a) Distribution of solutions

(b) Fitness of the best solution

Figure 2: Distribution of solutions around the peaks and the fitness of the best solution in the population for a 10-peak problem using a population size of 100.

we want to make. We are not here advocating that one particular crossover operator is better than another crossover operator, nor that it is better than mutation. What we want emphasize here is that when a GA solves the 2-peak problem, it fairly quickly focuses the population on a single peak, and from that point on the problem is akin to the trivial onemax problem.

Notice also, that in this particular run, the population did focus its attention to the very best peak. The reason why that happened was because the population was sufficiently large in order to decide, in a statistical sense, in favor of the best peak.

Now let us do the same experiments for a 10-peak problem. Using the same parameter settings as before, we did a single run on a single instance of a 10-peak problem. The results are depicted in figures 2. The behavior of the GA on this instance is identical to the one obtained with the 2-peak problem. Again, around generation 10, the whole population is focused around a single peak, which happen to be again the best peak. Once the population was focused around that peak, the GA proceeded to solve it very easily, again just like onemax.

We can now go further and test the GA with the exact same settings on an instance of a 100-peak problem. In this case it would be hard to read a plot with 100 lines showing the distribution of solutions around peaks through time. Rather than doing that, we use a bar plot for showing the distribution of solutions at particular points in time. Specifically, parts a), b), and c) of Figure 3, show the distribution of solutions at generation 0, 5, and 10, respectively. By generation 16, the whole population is at the basin of attraction of a single peak. This time however, it was not the best peak (no need for figure at generation 16—it is a single bar for peak 7). Figure 3-d shows the fitness of the best population member through time. The algorithm reaches a maximum fitness value of 0.965 by generation 70 (which correspond to the height of peak 7), and then fluctuates around that value until the end of the run.

This time, it took slightly longer for the GA to settle around a single peak. But by generation 16, all the population members were near peak 7. Again, once that happened the GA proceeded to solve the problem, climbing that peak just like it does on onemax. This time, however, it didn't reach the very best peak, and because of that, the GA continued to run until the 30 thousand function evaluations elapse (300 generations).[1] But it should be obvious to the reader that even if we had let the GA run for a much longer time, it would be nearly impossible for the algorithm to reach the very best peak. We say nearly impossible because when mutation is turned on, it is always possible to reach any other solution in the search space. But it is easy to recognize that in this case it would take an exponential time to do so. The reason is simple. Once the population members are all concentrated at the basin of attraction of the same peak, it is just a matter of a few generations until all the population members become very close to the top of that peak. At that point, recombination will not be able to do much other than generating two solutions near that same peak, and mutation will only be able to move a solution to the basin of attraction of another peak if something like $O(L)$ bits are mutated at once. Notice also that having a single solution at the basin of attraction of the best peak would not be sufficient. It would have to be a solution very close to the top of the best peak, or otherwise it would not be propagated by selection. Given a random layout of peaks, the expected hamming distance between any two peaks is $L/2$ bits. In order for mutation to move a solution from the top of some peak to the top of the best peak, it would have to mutate on average $L/2$ bits and not mutate the remaining $L/2$ bits. With a mutation rate of $1/L$ (or with any other mutation rate) it would take an exponential time to do such a jump!

---

[1] For the sake of simplicity, we evaluate the whole population in each generation, but we are aware that in a careful implementation we could evaluate only those individuals which have been affected by crossover or mutation

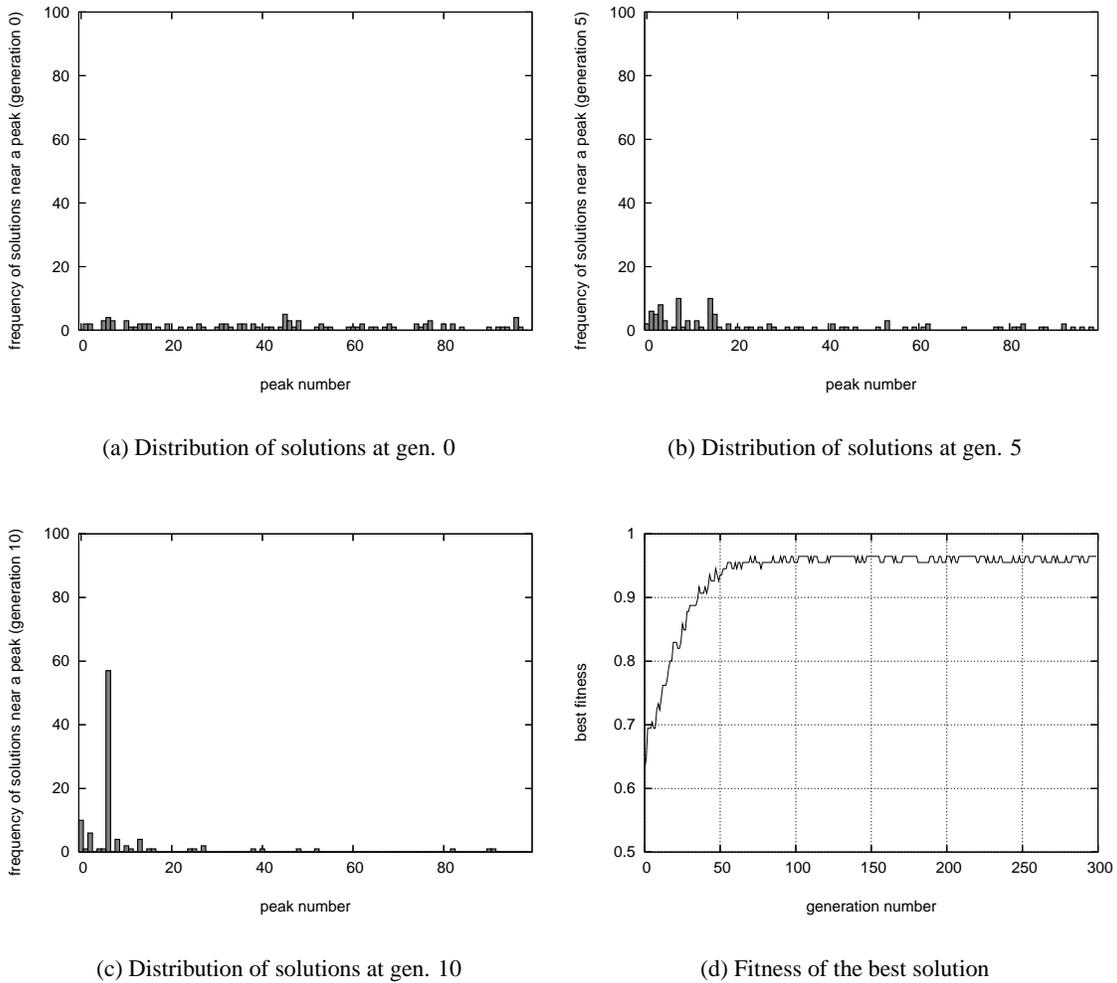

(a) Distribution of solutions at gen. 0

(b) Distribution of solutions at gen. 5

(c) Distribution of solutions at gen. 10

(d) Fitness of the best solution

Figure 3: Distribution of solutions around the peaks at generation (a) zero, (b) five, (c) ten, and the (d) fitness of the best solution in the population for a 100-peak problem using a population size of 100.

We now set forth an hypothesis that if we increase the population size, the GA will have a better chance at focusing on the best peak, and therefore reliably solve the problem to optimality. To test our hypothesis we run the GA with the exact same settings but varying the population size with exponentially increasing sizes: 100, 200, 400, 800, 1600, 3200, 6400, 12800, 25600, and 51200. For each population size we conduct 100 independent runs. We let the algorithm run until either it finds the best solution, or the distribution of the population members is concentrated around a peak other than the best one (once that happens it takes exponential time to solve the problem to optimality and we don't have time to wait for that). Figure 4 shows the success rate (SR), the number of runs that were able to reach the best peak for the different population sizes, as well as the average number of function evaluations (AES) needed by the algorithm to reach the best peak. This measure is an average of only those runs which were successful in finding the best peak.

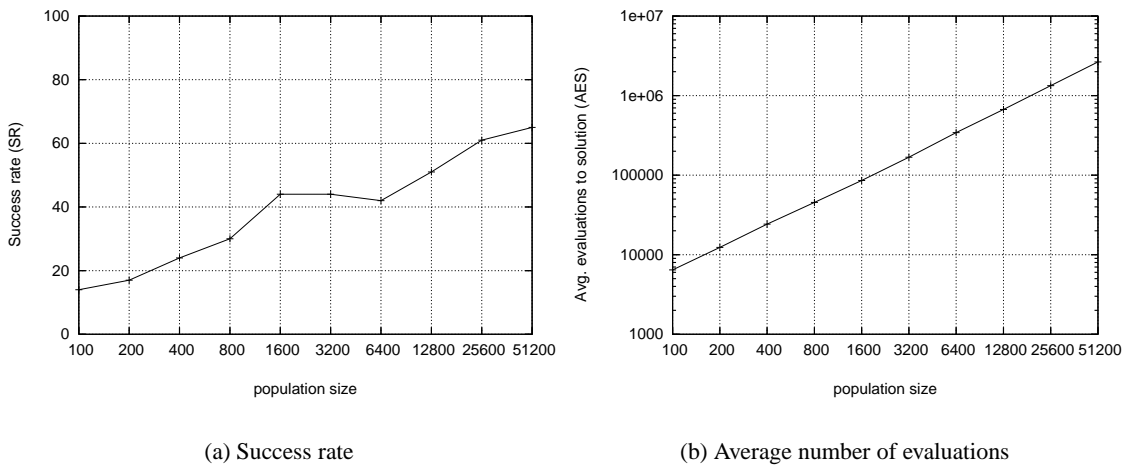

(a) Success rate    (b) Average number of evaluations

Figure 4: Success rate and average number of evaluations to find the highest peak when solving an instance of a 100-peak problem.

Notice that increasing the population size does help the GA to obtain higher success rates, but the population sizing requirements are large. Even a population size of 51200 was only sufficient to reach a 65% success rate, and we did not do further experiments with larger sizes. Nonetheless, it can be expected that by raising the population size even further we can get closer and closer to a 100% success rate. The AES measure grows linearly with the population size. That makes sense because the number of generations needed to climb a peak is more or less independent of the population size (Mühlenbein & Schlierkamp-Voosen, 1993). Thus, doubling the population size makes the GA take twice as much function evaluations to reach the highest peak.

## 5  Are Evolutionary Algorithms adequate for these problems?

The results that we have been presenting so far, suggest that the relative merits of crossover over mutation and vice-versa is a secondary issue. We have shown that an EA rapidly concentrates its population members around a single peak, and from that point on, the problem becomes equivalent to the onemax problem, for which the debate of crossover versus mutation is rather futile since both can easily solve the problem. What our results suggest is that the population size plays a crucial role in the ability of an EA to reliably reach the highest peak. The population sizing requirements are large, and clearly depend on 2 issues: (1) we need to have enough samples of solutions at the basin of attraction of the best peak right from the beginning of the search, and (2) the EA has to be able to propagate those solutions rather than propagating solutions at the basin of attraction of other peaks. Although we did not derive a population sizing model for this class of problems, it is clear that such a model would have to take into account the number of peaks, and the ability of the selection operator to distinguish between solutions near the best peak, and solutions that are near the second best peak (the most tough competitors). As pointed out before, there is a strong connection between these issues and the building block supply and decision making issues from existing population sizing models on additively decomposable problems (Harik, Cantú-Paz, Goldberg, & Miller, 1999; Goldberg, Deb, & Clark, 1992).

Table 1: Multi-restart nest ascent hillclimbing on a 100-peak problem.

| Success rate | 100 |
|---|---|
| Avg. evaluations to solution | 22779 |
| Avg. number of restarts | 52.9 |

We now stop for a moment to reflect whether an EA is an appropriate algorithm to solve this class of problems. The answer is no and there's a simple reason to it. EAs are good when there is some structure to be exploited. Instances from the multimodal problem generator have no such structure. Peaks are random and have nothing to do with one another. Thus, in order to solve the problem the EA has to either get lucky and concentrate on the best peak, or use a population size that should be at least $O(\mathcal{P})$. But if those are the requirements for reliably solving the problem, then a multi-restart hillclimbing algorithm should be able to do the job much faster. Such an algorithm has no population at all. Starting from a random solution, it climbs up the peak. Once there, it restarts from another random solution and climbs the peak again, and so on until a specified maximum number of function evaluations has elapsed or some other stopping criterion is reached. On a problem with $\mathcal{P}$ peaks, we should expect an average of $\mathcal{P}/2$ restarts.

We have conducted such experiments using a multi-restart next-ascent hillclimbing algorithm. In next-ascent hillclimbing, the bits are flipped in a predefined (randomly generated) sequence. A flip is accepted if the new solution has a higher fitness than the current solution. In that case, the new solution becomes the current solution, and the process continues until no further improvement is possible by flipping a single bit.

We performed 100 independent runs of a multi-restart next-ascent algorithm, and for each one we let the algorithm run until either it found the best peak, or a maximum of 1 million function evaluations was reached. The results are summarized in table 5. As expected, the multi-restart next ascent algorithm consistently reaches the highest peak in all runs, taking on average close to 23 thousand evaluation, and needing on average 53 restarts to do so. We are tempted do say that solving a problem instance from the multimodal generator is not as difficult as finding a needle-in-a-haystack but it is really like finding a *peak-in-a-haystack*. To do such a job, it is unlikely that any EA is capable of doing any better than a multi-restart hillclimbing algorithm.

It should be pointed out that even iterated local search algorithms (Lourenco, Martin, & Stützle, 2002) would have difficulties in solving these problems unless they are capable of adjusting their perturbation strength to move very far away form an optima. In other words, to do a complete restart. The reason for this has been mentioned before. Peaks have nothing to do with one another, and that goes against the basic principles of operation of iterated local search methods. These methods, search on the space of local optima. The problem is that in this case, a local optima gives no information whatsoever as to whether other local optima might be.

## 6 Summary

This paper looked in detail at the multimodal problem generator, a test problem generator that has been used by many researchers to assess the performance of evolutionary algorithms.

After setting an hypothesis of how an EA attempts to solve problems from this class, we conducted com-puter simulations that confirmed our suspicion. It was shown that the EA rapidly concentrates its population around a single peak, and after that, the algorithm simply climbs that peak, having virtually no chance from that point on to reach another peak.

The paper then showed that the only way to reliably solve these problems with an EA is by raising the population size so that supply and decision making issues are taken into consideration. Finally, the paper argued that the multimodal problem generator produces instances that are like a peak-in-a-haystack, and that EAs are not the most adequate algorithms for solving such problems.

# 7 Conclusions

Several researchers have been using the multimodal problem generator for assessing and comparing the performance of various evolutionary algorithms. Based on those experimental studies, researchers have been making different claims as to the adequacy of an operator over another, or the relative merits of a particular EA with respect to another. The results presented in this paper suggest that such arguments are futile and that it really does not matter what operator is used as long as the population size is large enough. The paper shows that on this class of problems, the EA very quickly concentrates the population at the basin of attraction of a single peak, and once that happens, the problem becomes akin to the the onemax problem, which has been widely studied by the evolutionary computation community.

Finally, the paper argues that the multimodal problem generator should not be used for assessing the performance of EAs because EAs are simply not the most adequate problem solvers for such a class of problems. Due to the characteristics of the problem generator, it is very unlikely that any EA is capable of beating (both in speed and reliability) the performance of a multi-restart hillclimbing algorithm.

Notice that we are not saying that the multimodal problem generator is uninteresting. It is in fact quite instructive in that it reveals interesting algorithm dynamics, and reveals the importance of properly sizing the population of an evolutionary algorithm, something that continues to be underestimated by a large number of researchers in the field.

We also do not make any claims regarding the existence or not of real world problems with characteristics similar to those of instances obtained from the multimodal generator. But one thing is sure. If there are real world problems with such characteristics, then an evolutionary algorithm (no matter if it has this or that operator) is certainly not the algorithm of choice.

As a final remark, we would like to say that the idea of having problem generators is, in our opinion, a nice idea and an interesting research topic to pursue as far as experimental research is concerned. But in order to be useful for assessing the performance of EAs, it is important that the generators are capable of generating problem instances that have some sort of structure that can be exploited by EAs, as opposed to problem instances that are relatively easily solvable by hillclimbing alike algorithms.

# Acknowledgments

The authors thank the support of the Portuguese Foundation for Science and Technology (FCT/MCES) under grants POSI/SRI/42065/2001, POSC/EEA-ESE/61218/2004, and SFRH/BD/16980/2004.